\documentclass[10pt,twocolumn,letterpaper]{article}

\usepackage{wacv}              

\usepackage{graphicx}
\usepackage{amsmath}
\usepackage{amssymb}
\usepackage{booktabs}
\usepackage{multirow}
\usepackage{rotating}
\usepackage{newfloat}
\usepackage{listings}

\usepackage{boldline}

\usepackage{algorithm}
\usepackage[noend]{algpseudocode}
\usepackage{rotating}
\usepackage[table,xcdraw]{xcolor}
\usepackage{xcolor, soul}
\usepackage{makecell}
\usepackage[table]{xcolor}
\usepackage[accsupp]{axessibility} 

\usepackage[pagebackref,breaklinks,colorlinks]{hyperref}
\usepackage{color,array}
\usepackage{graphicx}
\usepackage{booktabs}
\usepackage{colortbl}
\usepackage[capitalize]{cleveref}
\crefname{section}{Sec.}{Secs.}
\Crefname{section}{Section}{Sections}
\Crefname{table}{Table}{Tables}
\crefname{table}{Tab.}{Tabs.}

\usepackage[accsupp]{axessibility}  

\def\wacvPaperID{2653} 
\def\confName{WACV}
\def\confYear{2025}

\begin{document}

\title{Phaseformer: Phase-based Attention Mechanism for Underwater Image Restoration and Beyond}
\author{
    MD Raqib Khan$^{1}$, Anshul Negi$^{2}$, Ashutosh Kulkarni$^{2}$, Shruti S. Phutke$^{2}$, \\
    Santosh Kumar Vipparthi$^{2}$, Subrahmanyam Murala$^{1}$\\
    $^{1}$CVPR Lab, School of Computer Science and Statistics, Trinity College Dublin, Ireland\\
    $^{2}$CVPR Lab, Indian Institute of Technology Ropar, India\\
    {\tt\small khanmd@tcd.ie}  
}

\maketitle

\begin{abstract}
\vspace{-3mm}
Quality degradation is observed in underwater images due to the effects of light refraction and absorption by water, leading to issues like color cast, haziness, and limited visibility. This degradation negatively affects the performance of autonomous underwater vehicles used in marine applications. To address these challenges, we propose a lightweight phase-based transformer network with 1.77M parameters for underwater image restoration (UIR). Our approach focuses on effectively extracting non-contaminated features using a phase-based self-attention mechanism. We also introduce an optimized phase attention block to restore structural information by propagating prominent attentive features from the input. We evaluate our method on both synthetic (UIEB, UFO-120) and real-world (UIEB, U45, UCCS, SQUID) underwater image datasets. Additionally, we demonstrate its effectiveness for low-light image enhancement using the LOL dataset. Through extensive ablation studies and comparative analysis, it is clear that the proposed approach outperforms existing state-of-the-art (SOTA) methods. Code is
available at \href{https://github.com/Mdraqibkhan/Phaseformer}{ Phaseformer}.


\vspace{-3mm}
\end{abstract}

\vspace{-3mm}

\section{Introduction}
\vspace{-2mm}
\begin{figure}[htbp]
\begin{center}
  \includegraphics[width=1\linewidth]{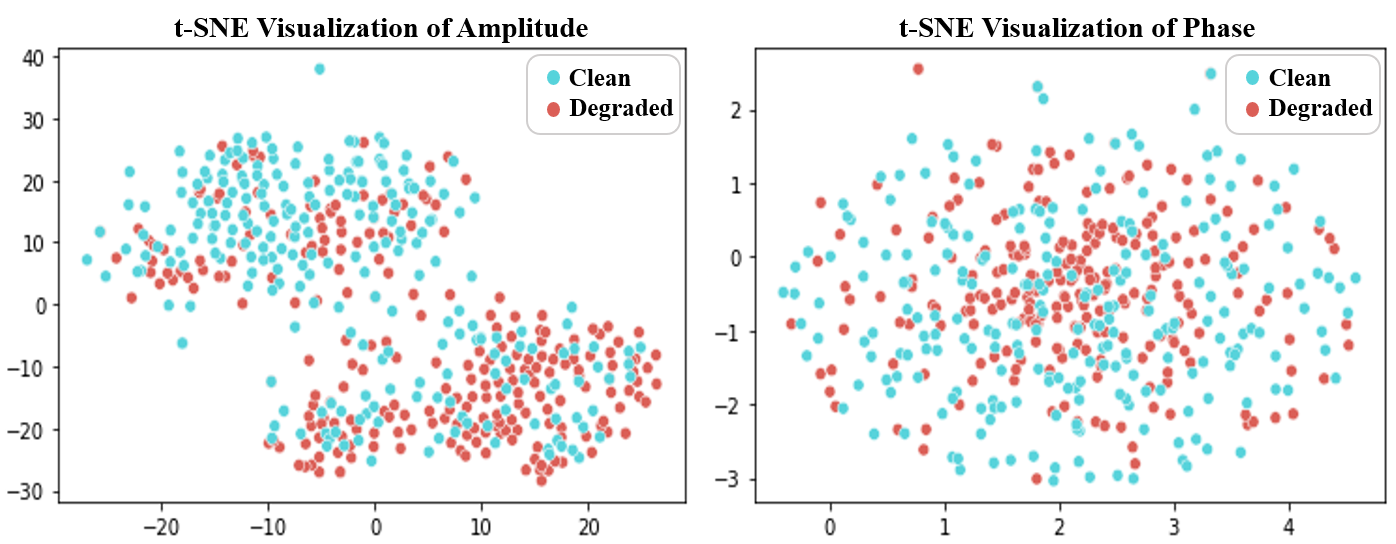}
\end{center}
\vspace{-4mm}
   \caption{t-SNE visualization of the Amplitude and Phase of clean and degraded images. The separate clusters for clean and degraded amplitude show that there is more effect of degradation on amplitude content as compared to phase content which has overlapping clusters for clean and noisy images.}
   \vspace{-4mm}
\label{fig:tsne}
\end{figure}
Underwater images with higher perceptual quality have a higher significance in marine engineering \cite{liu2020real} and aquatic robotics. However, effects like light scattering may degrade the quality of underwater images, including blurriness, reduced brightness, low contrast, and haziness. These degradations significantly affect the performance of underwater applications. To mitigate these challenges, underwater image restoration (UIR) is a feasible solution to enhance the perceptual quality of underwater images, increasing their visibility which potentially leads to improved accuracy in automated underwater applications.
Therefore,  various researchers proposed algorithms to handle these degradations \cite{williams2010optimal,li2018emerging,zhang2022underwater,henderson2013mapping,caimi2008underwater,drap2012underwater,liu2020real}.

\par The prevailing works towards UIE can be classified into three categories. First, the methods following the physical model for estimating the transmission map to generate enhanced images, but it may struggle in complex and dynamic underwater environments \cite{he2010single,drews2013transmission,wang2017single,peng2017underwater,li2016underwater}. Second, the visual prior-based approaches focus on enhancing visual quality but ignore the physical deterioration process \cite{li2016underwater,abdul2014underwater,fu2014retinex,huang2018shallow,iqbal2010enhancing}. Last, the deep-learning methods, including transformers, have shown impressive performance in various image restoration tasks, including UIE \cite{guo2022image,valanarasu2022transweather,tsai2022stripformer,Zamir2021Restormer,ren2022reinforced,peng2023u}. Since UIE is a pre-processing task, it is of utmost importance for the UIE architecture to be less computationally complex. Motivated by this, we propose a lightweight architecture for underwater image restoration. 
\par During the underwater image acquisition process, the impact of degradation is greater on the amplitude of image than on the phase \cite{haykin2008communication,khan2023underwater}. Figure \ref{fig:tsne} shows the t-SNE visualization of the amplitude and phase of degraded and clean underwater images. This visualization effectively highlights the disparities in the distribution of phase and amplitude between the two sets and offers valuable insights into the impact of degradation on image quality. The separate clusters of amplitude from clean and degraded images show that there is a change in amplitude information of an image when it is degraded. Whereas, the phase of clean and degraded images have overlapping clusters showing less effect of degradation on it. Since the phase of an image represents structural information \cite{1456290}, it becomes handy to utilize the phase information to enhance the degraded underwater image. With this motivation of phase being less sensitive to degradation, we propose a phase-based self-attention mechanism in the transformer block to achieve efficient color restoration.
\par The transformer-based U-Net architectures \cite{ghani2015underwater} for underwater applications \cite{jaffe1990computer,ren2022reinforced} solve the vanishing gradient problem via skip connection between the encoder to the respective decoder. The direct skip connection from the encoder to the decoder may pass even some redundant content. Further, as depth increases the higher wave-length colors get attenuated at a higher rate as well as there may be variations in structural information also. The phase represents the structural information of an image that is less sensitive to degradation (see Figure \ref{fig:tsne}). In light of this, for structural detail propagation from encoder levels to decoder, instead of using direct skip connections, we propose a phase attention block. \textit{To the best of the authors' knowledge, this is the first attempt focused on utilizing the phase-based transformer mechanism for underwater degraded image restoration.}

Also, the proposed network utilizes various losses and the weight of each loss plays an important role to gain optimized results. But manual weight tuning becomes a tedious task. Therefore to solve this issue, we propose an adaptive weight assignment technique that automatically tunes the weights of different losses. Overall, the main contributions of our work are:
\begin{itemize}
\item A novel phase-based lightweight transformer (Phaseformer) architecture is proposed for underwater image restoration.
\item A novel phase-based self-attention mechanism is introduced in the transformer, that operates on the prominent (phase) information of images, enabling it to gather both local and non-local prominent pixel interactions.
\item An optimized phase attention block is proposed to forward the attentive structural information from the encoder to the respective decoder for effective restoration.
\item An adaptive weight assignment to the loss functions is proposed while training the network for better optimization.
\end{itemize}
The ablation study is done on different configurations of the proposed approach. We carry out several experiments for demonstrating effectiveness of the proposed method on both synthetic as well as real-world images for underwater image enhancement. Also, we verify the applicability of the proposed method for low-light enhancement task.


\section{Related Work}
\vspace{-2mm}
\begin{figure*}[htbp]
\begin{center}
  \includegraphics[width=1\linewidth]{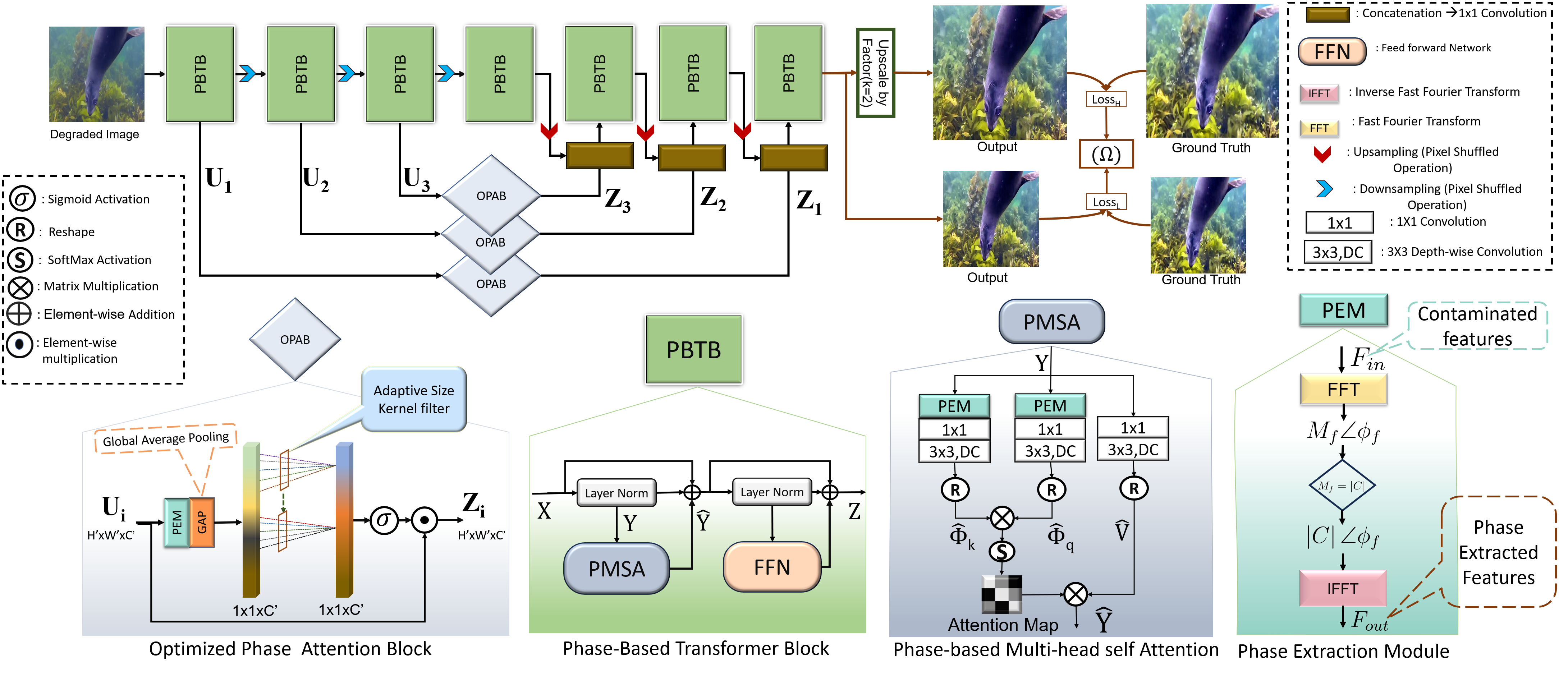}
\end{center}
\vspace{-5mm}
   \caption{Architectural schematic of the proposed underwater image enhancement network. The network comprises of \textbf{Phase-based Transformer Block, Optimized Phase Attention Block, and dynamic weight assignment to the loss functions}. The proposed phase-based transformer block captures local and non-local structural information using less attenuated phase-based queries and keys. Optimized Phase Attention block is proposed for propagating prominent attentive features from the encoder to the respective decoder. The dynamic weight ($\Omega$) assignment is proposed for efficient optimization of the network while training.}
\label{fig:Network}
\end{figure*}
\vspace{-2mm}

\par Initial attempts towards enhancing the underwater images mainly relied on varying the pixel intensities in the red and the blue channels. \cite{ghani2015underwater} followed the Rayleigh distribution for obtaining enhanced underwater image by shifting the blue channel portion, while concurrently positioning the red channel in the upper portion. The Jaffe-McGlamery model has served as a foundation for transmission estimation techniques in research \cite{jaffe1990computer}. Following this model, \cite{liu2016underwater} optimized the transmission map using a quadtree-based loss function to enhance contrast in the images. \cite{li2017hybrid} estimated the medium transmission via a combination of regression modeling and a global underwater light estimator. For visual result enhancement, \cite{chiang2011underwater} proposed a combination of the wavelength-dependent compensation along with the dark channel prior combined with the dark channel prior.  
 \cite{peng2018generalization} proposed an approach involving a generalized dark channel prior, incorporating adaptive color correction for underwater images. However, this method leads to overexposure and incomplete effectiveness in addressing various underwater challenges. \cite{li2019underwater} extended CNNs with the introduction of the UIEB dataset and Water-Net for UIE. 
 \cite{li2020underwater} employed synthesized underwater images to train their proposed UWCNN, simulating a realistic underwater environment for image enhancement. \cite{jamadandi2019exemplar} introduced a wavelet-corrected transformation for enhancing underwater images.\cite{guo2019underwater} proposed a weakly supervised learning based generative adversarial network, incorporating multiple scaled processing of features. \cite{li2021underwater} proposed a multi-color processing network for representing a diverse range of color features. For simultaneously enhancing the performance of image enhancement and object detection, \cite{chen2020perceptual} introduced dual perceptual enhancement modules for detection, recognizing the need to balance image quality optimization with improved detection outcomes. By analysing the light absorption and blurriness, \cite{peng2017underwater} estimated depth maps for enhancing the visibility in underwater images. Few works
  \cite{islam2019fast,fabbri2018enhancing} utilized conditional adversarial networks (CGAN) for enhancing underwater images. \cite{sharma2021wavelengthbased} introduced a unified underwater image enhancement and super-resolution approach based on multi-receptive attention. \cite{liu2022twin} proposed a twin adversarial contrastive learning-based method that further leverages detection-aware training for enhancing the underwater images.

  \par Transformers are widely used in various image restoration tasks. Applications include image dehazing with a hybrid CNN-Transformer \cite{guo2022image}, efficient image deraining, denoising, and deblurring \cite{Zamir2021Restormer}, as well as the ``Swin Transformer" \cite{liu2021swin}, for computational efficiency in image denoising and deblurring. Additionally, a U-shaped transformer is used for UIE with feature fusion modules \cite{peng2023u}.

\section{Proposed Method}
\vspace{-2mm}
The existing approaches generally utilize the different color spaces as inputs with separate encoder-decoders
to process the original RGB and different color spaces for UIR, which may increase the computational cost of the overall architecture. To mitigate such limitation, here, we propose a lightweight architecture for UIR named "Phaseformer: phase-based attention transformer network". For considering the peculiarities of underwater imaging, we propose phase-based self-attention and optimized phase attention. In this paper, we initially illustrate the complete pipeline of our architecture, as shown in Figure \ref{fig:Network}. Afterward, we provide a detailed overview of the proposed phase-based self-attention and optimized phase attention.
\newline
\textbf{Overall Pipeline:} Initially, the input image is passed through a series of proposed phase-based transformer blocks (PBTB) consisting of phase-based self-attention. The features from the initial PBTB are propagated through the proposed optimized phase attention utilized in skip connections. At the final stage, we obtain the outputs with two different resolutions named as actual resolution and  $\times2$ resolution by applying transposed convolution. These multi-resolution outputs are further utilized for loss calculations (see Figure \ref{fig:Network}). The proposed sub-modules are discussed below:
\vspace{-3mm}

\subsection{Phase-based Transformer Block}
\vspace{-1mm}
Studies show that the phase of an image contains the most relevant information \cite{juvells1991role,oppenheim1981importance,Xu_2021_CVPR,hansen2007structural,NEURIPS2018_647bba34,park2018adversarial,oppenheim1979phase}, which is sufficient to reconstruct the image completely.
Further, it has been observed that the phase of an image is less affected by noise compared to its magnitude \cite{chan2005image,haykin2008communication}. 
The t-SNE plot depicted in Figure \ref{fig:tsne} illustrates the comparison between the phase and amplitude of both clean and degraded images. This visualization effectively highlights the disparities in the distribution of phase and amplitude between the two sets and offers valuable insights into the impact of degradation on image quality. As a result, many image processing algorithms place significant emphasis on phase information.
\par In traditional self-attention mechanism, most of the methods \cite{46840,Wang2019,Beltagy2020Longformer,wang2020linformer,DBLP:journals/corr/abs-2103-02143} have a time complexity that grows linearly as the input patch size increases. The attention map is generated by computing the dot product of the query and key vectors and applying a softmax function to obtain a probability distribution over the values. Resulting attention map is then used to weigh the value vectors and compute the output of the attention mechanism. However, this approach may not be sufficient for image restoration tasks that require fine-grained spatial relationships. The phase of an image provides valuable spatial information \cite{juvells1991role,oppenheim1981importance,Xu_2021_CVPR,hansen2007structural,NEURIPS2018_647bba34,park2018adversarial,oppenheim1979phase}. Incorporating this information into the self-attention mechanism may help to generate more fine-grained and interpretable attention maps that capture the spatial relationships between different parts of the image. With this motivation, we propose a phase-based self-attention block, shown in Figure \ref{fig:Network} that takes structural information through phase features as key and query to generate a more accurate and effective attention map.

 In the proposed phase-based transformer block (PBTB), first, the normalized tensor $Y \in \mathbb{R}^{H' \times W' \times C'}$ is fed to the proposed phase-based Multi-head self-attention (PMSA) mechanism (see PBTB in Figure \ref{fig:Network}). In PMSA, the to generate the structural features for key and query, we proposed a phase extraction module (PEM) (see PMSA in Figure \ref{fig:Network}). For this, we applied the FFT on input features to extract the phase ($\phi_f$) and amplitude ($M_f$) information. Further, by making $M_f = C$ ($C=1$) and keeping the phase ($\phi_f$) as it is, the IFFT is calculated (see PEM in Figure \ref{fig:Network}). This operation provides us the prominent phase-only information from the input feature maps which is then fed to the multi-head attention as query ($\Phi _{q}$) and key($\Phi _{k}$).
 
On these phase-query  ($\Phi _{q}$), phase-key ($\Phi _{k}$) from PEM, and value ($V$) from input projection the multi-head attention is applied, which enrich the local context. Subsequently, both query and key projections are reshaped such that their dot-product results in a transposed-attention map ($Attention$) of size $\mathbb{R}^{C' \times C'}$.
\vspace{-2mm}
\begin{equation}
  \hat{\Phi}_q = \varphi_3(\psi_1(PEM(Y))); \,\,\, \hat{\Phi}_k = \varphi_3(\psi_1(PEM(Y)))
\end{equation}
\vspace{-2mm}
\begin{equation}
\hat{Y} =  \psi_1\left( \text{Attention} \left(\hat{\Phi}_{q}, \hat{\Phi}_{k}, \hat{V}\right)\right) + Y
\end{equation} 
\vspace{-2mm}
\begin{equation}
 \text{Attention} \left(\hat{\Phi}_{q}, \hat{\Phi}_{k}, \hat{V}\right) = \hat{V} \cdot \text{Softmax} \left(\frac{\hat{\Phi}_k \cdot \hat{\Phi}_q}{\alpha}\right)
\end{equation}  
where, $Y$ and $\hat{Y}$ are the input and output feature maps of the proposed PMSA. For aggregating pixel-wise cross-channel context, a convolution operator ($\psi_m(\cdot)$) of kernel size ($m\times m$) is used. For channel-wise spatial context, depth-wise convolution operator ($\varphi_m(\cdot)$) of kernel size ($m\times m$) is used, where $m\in (1,2,3)$. The network uses bias-free convolution layers.  $\hat{\Phi}_q \in \mathbb{R}^{H'W' \times C'}$, $\hat{\Phi}_k \in \mathbb{R}^{C' \times H'W'}$, and $\hat{V} \in \mathbb{R}^{H'W' \times C'}$ matrices are obtained after reshaping tensors from the original size $\mathbb{R}^{H' \times W' \times C'}$. Parameter $\alpha$ can be learned to adjust the dot product of $\hat{\Phi}_q$ and $\hat{\Phi}_k$ before the application of the softmax function. This helps in controlling the magnitude of the dot product. We follow a similar approach as conventional multi-head self-attention (SA) \cite{dosovitskiy2020image}, where we split the channels among separate ``heads" and learn distinct attention maps in parallel. Further, the $\hat{Y}$ is processed through a feed-forward network (FFN) \cite{zamir2022restormer} to get the output of PBTB ($Z$ as shown in PBTB block of Figure \ref{fig:Network}). Processing the input values with phase based attention helps to restore color information effectively.


\subsection{Optimized Phase Attention Block}
\vspace{-1mm}
In general, in order to assist in the reconstruction process skip connections are utilized to forward the encoder features to the corresponding decoder features \cite{ronneberger2015u}. Directly forwarding the features as skip connection may result in some degree of degradation being transferred to the decoder, thereby hindering its ability to efficiently produce an enhanced image.
The existing work \cite{peng2023u} employed channel-wise multi-scale feature fusion transformer attention to transfer the attentive features from the encoder to the corresponding decoder. However, these attention blocks are computationally expensive and may forward degraded features to the decoder since they are processed in the spatial domain. Moreover, it has been observed that the structural information conveyed through the phase of an image is more resilient to degradation than the amplitude \cite{chan2005image,haykin2008communication}. This property may help with effective image enhancement when only phase information of encoder features is utilized. Therefore, we have introduced a lightweight optimized phase attention block (OPAB), depicted in Figure \ref{fig:Network}, which generates prominent and attentive features $Z_i$ as:
\vspace{-3mm}
 \begin{equation}
      Z_{i} =U_{i} \otimes \sigma(\omega _k(GAP(PEM(U_{i}))))
\end{equation} 
where, $U_{i}$ are the input features  of dimension $ \frac{H}{2^{i-1}} \times \frac{W}{2^{i-1}} \times 2^{i-1}C$, $i \in(1,2,3)$. These, $U_i$ are used to generate phase information by applying a phase extracting module (PEM).  Global average pooling operation (GAP) generates the spatially aggregated features with size $1\times 1\times 2^{i-1}C$. The $\omega _k$ is a 1D convolution operator with adaptive kernel size. In traditional CNNs, the kernel size is fixed and does not change during the training process. However, different encoder features may require different kernel sizes to be effectively captured by the network. This means that some features may be over-smoothed (due to large kernel size) or under-smoothed (due to small kernel size) by the fixed kernel size \cite{agrawal2020using}, resulting in loss of important information and hence reduced performance. OPAB addresses this issue by adaptively selecting the kernel size based on the number of input feature channels. It does this by applying a learnable 1D convolution layer to the encoder features, which is then used to weigh the features at each channel. This allows the network to learn which kernel size is best suited to capture the features in each channel of the input. The adaptive kernel size $k$ is determined by:
\vspace{-2mm}
\begin{equation}
    k=\wp \left ( C' \right )=\left| \frac{log_2(C')}{\gamma} + \frac{b}{\gamma}\right|_{odd}
\end{equation}
where, $C'=2^{i-1}C$ is the number of channels after GAP, $|t|_{\text{odd}}$ indicates the nearest odd number of $t$. In this paper, we set $\gamma$ and $b$ to 2 and 1 respectively.



 \subsection{Adaptive Weight Assignment for Training Losses}
 \vspace{-2mm}
 In deep learning architectures, it is common to train models using multiple loss functions to achieve various objectives simultaneously. However, these loss functions may have different scales or importance levels, which may lead to difficulty in optimizing the model. To address this issue, we proposed a learnable weight assignment technique, that enables the model to achieve better performance by combining multiple loss functions effectively. It provides a flexible and effective way to balance the trade-offs between different objectives, making it a valuable tool in various applications, such as image restoration.
 
 \begin{equation}
 L_{T}= \Omega_{H}*L_{H}+\Omega_{L}*L_{L}
 \end{equation}
 \begin{equation}
L_{R}= \Omega_{1}*L_{C}^{R}+\Omega_{2}*L_{G}^{R}+\Omega_{3}*L_{M}^{R}+\Omega_{4}*L_{P}^{R}
 \end{equation} 
 $L_R$ represents the total loss with $R \in (L,H)$ (L is low-resolution, H is high-resolution). The adaptive weights ($\Omega_i$, $i \in \{1,2,3,4\}$) are trainable tensors that are updated during training. These adaptive weights are applied to different loss functions, namely Charbonnier loss ($L_{C}$) \cite{bruhn2005lucas}, Gradient loss ($L_{G}$) \cite{ribeiro1995case}, Multi-Scale SSIM (MS-SSIM) loss ($L_{M}$) \cite{wang2003multiscale}, and Perceptual loss ($L_{P}$) \cite{johnson2016perceptual}. At the final epoch, the values of these weights are 0.2741, 0.2222, 0.3357, and 0.1680, respectively. Additionally, empirically determined weights are assigned to the high-resolution loss ($L_H$) and the low-resolution loss ($L_L$) as $\Omega_{H}= 0.4$ and $\Omega_{L}=0.6$, respectively. \textit{Comprehensive details regarding the employed loss functions are provided in the supplementary material.}
 \vspace{-3mm}
 \newline

\begin{table*}[h]
\centering
\caption{Quantitative comparison of the proposed method (Ours) with existing SOTA methods for underwater image restoration. ($\uparrow$) higher is better, \textbf{bold} and \underline{underline} indicate the \textbf{best} and \underline{second best} results.}
\vspace{-2mm}
\begin{tabular}{l|c|c|c|c|c|c|c|c}
\hline
 \rowcolor{gray!20}\multicolumn{9}{c}{\textbf{UIEB}} \\
\hline
Method& RGHS &WaterNet&UIECL &U-shape &TACL&Semi-UIR& Spectrofomer& \textbf{Ours}\\
\hline
PSNR$\uparrow$  & 14.57&  19.81 & 20.37 &22.91 &23.72& {24.59}&\underline{24.96}& \textbf{25.98} \\
    SSIM$\uparrow$ & 15.78 & 0.731 & 0.864 & {0.910} & 0.830 &0.901&\underline{0.917} &\textbf{0.928}\\
UIQM$\uparrow$  &3.047&3.836 & 3.544&3.726 &3.969 &\textbf{4.598}& 3.075&\underline{4.451}\\
\hline
\rowcolor{gray!20}\multicolumn{9}{c}{\textbf{UFO-120}} \\
\hline
Method  &UDCP&UIBLA &RGHS&FGAN&FGAN-UP & Deep-Sesr&UIECL&\textbf{Ours}\\
\hline
PSNR$\uparrow$       &16.05 & 20.34 &21.03 & 25.15 &  24.89  & \underline{27.15}  & 20.41  & \textbf{30.38} \\
SSIM $\uparrow$     & 0.571 & 0.657   & 0.739  & 0.820 &0.759& \underline{0.840}  & 0.754 &  \textbf{0.878} \\
UIQM$\uparrow$ & 1.836 &1.99  &2.271 & 2.668 & 2.597& \underline{2.749} & 2.703 &  \textbf{2.907} \\
\hline
\end{tabular}
\label{tab:uieb}
\end{table*}


\begin{figure*}
\begin{center}
  \includegraphics[width=1\linewidth]{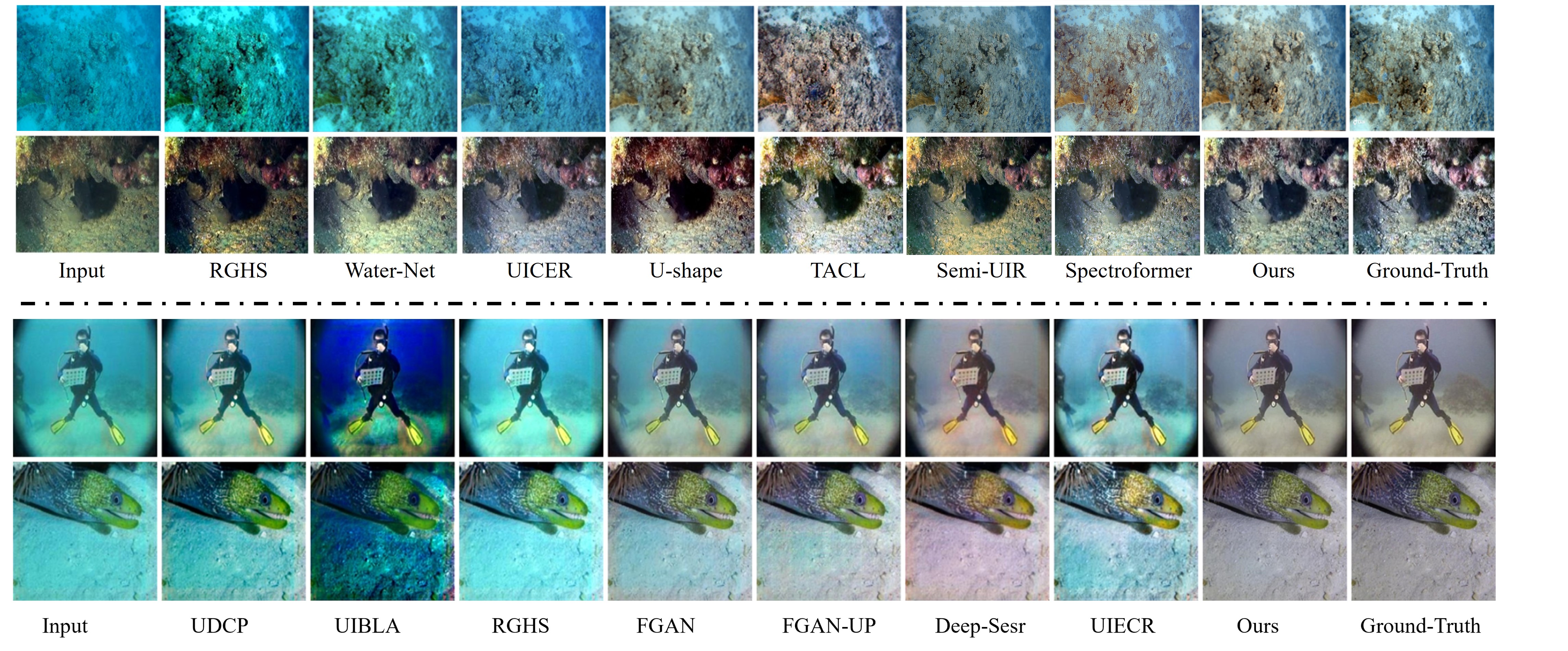}
\end{center}
\vspace{-5mm}
   \caption{Qualitative comparison of the proposed method (Ours) with existing SOTA methods for underwater image restoration on the synthetic datasets (upper part: UIEB dataset, lower part: UFO-120).}
   \vspace{-1mm}
\label{fig:uieb_result}
\end{figure*}

\begin{figure*}[t]
\begin{center}
  \includegraphics[width=1\linewidth]{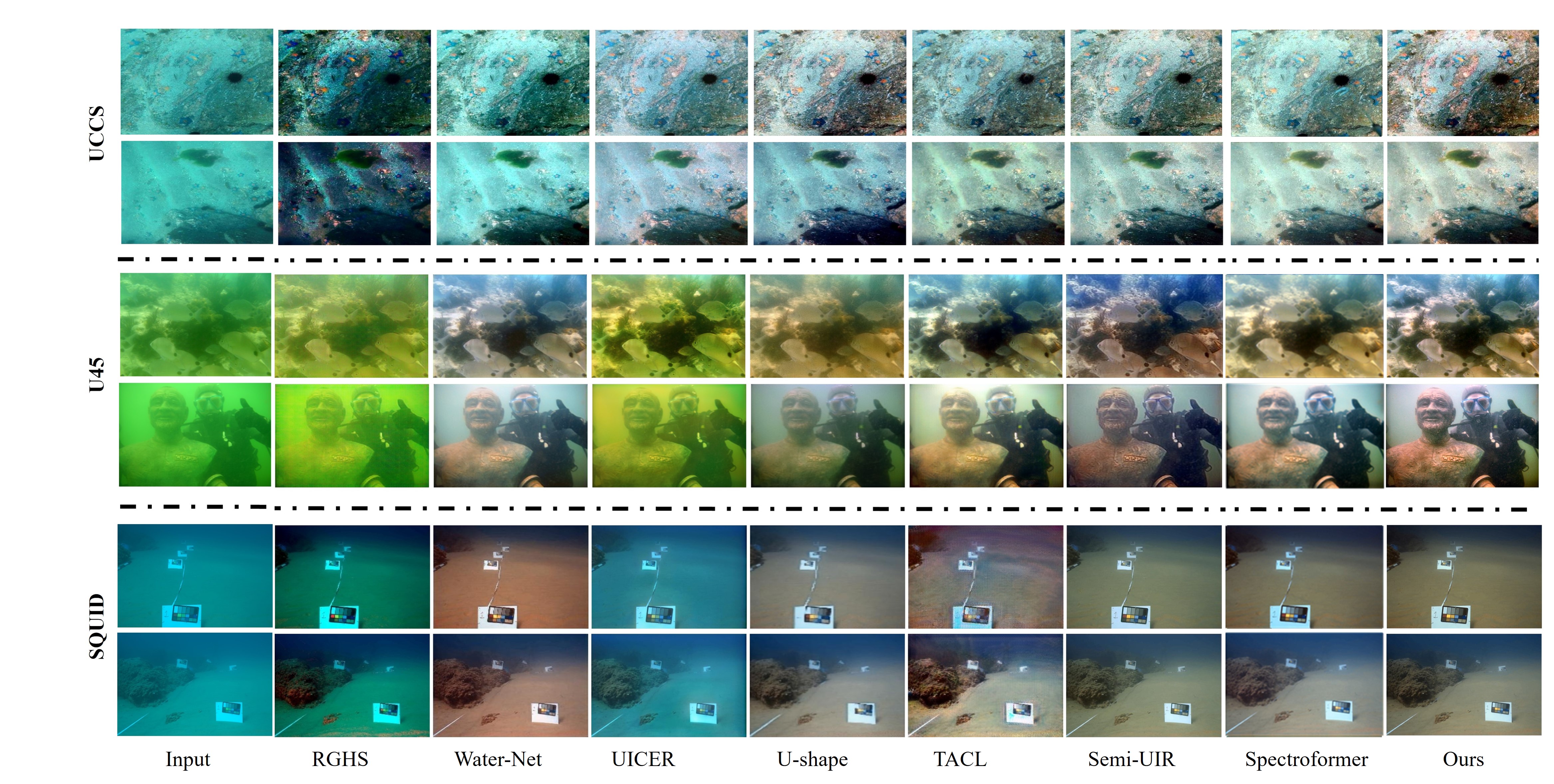}
\end{center}
\vspace{-4mm}
   \caption{Qualitative comparison of the proposed method (Ours) with existing SOTA methods for underwater image restoration on real-world UCCS, U45, and SQUID datasets.}
\label{fig:euvp_result}
\vspace{-2mm}
\end{figure*}

\section{Experimental Discussion}
\vspace{-1mm}
\subsection{\textbf{Datasets}} 
\vspace{-2mm}
To conduct a comparative analysis, we have considered synthetic underwater UIEB\cite{li2019underwater}, UFO-120 \cite{islam2020simultaneous} and real-world underwater U45 \cite{li2019fusion}, UCCS\cite{liu2020real}, and SQUID \cite{berman2018single} datasets. The underwater image enhancement benchmark (UIEB) dataset comprises 890 underwater images, featuring various scenes to ensure a wide range of conditions. The training set is composed of randomly selected 800 image pairs, while the remaining 90 images are considered for testing purpose.
UFO-120 contains 1500 pairs in training set and a test set of 120 pair samples. U45 comprises 45 real-world images that showcase characteristics such as color casts, low contrast, and the degradation effects resembling haze in underwater scenarios. 
The SQUID dataset comprises 57 sets of stereo pairs captured at various locations within Israel.

\subsection{\textbf{Training Details}} 
\vspace{-2mm}
For generating the necessary training images, we applied data augmentation methods, including horizontal and vertical flipping, noise addition, and contrast variation. This led to the creation of 3000/4800 training image pairs and 120/90 testing image pairs from the UFO-120/UIEB datasets, respectively. All input images were resized to $256\times 256$ dimensions. During training, we utilized the ADAM optimizer with initial learning rate of $3\times 10^{-4}$, which was adjusted using the cosine annealing strategy. The implementation of the proposed network was carried out using PyTorch and training was performed on an NVIDIA GeForce RTX 2080 GPU.
\subsection{\textbf{Baselines}} 
\vspace{-2mm}
For quantitative and qualitative comparison, we have considered UDCP \cite{drews2016underwater}, UIBLA \cite{peng2017underwater}, RGHS \cite{islam2020fast}, WaterNet \cite{li2019underwater}, FGAN \cite{islam2020fast}, FGAN-UP \cite{islam2020fast}, Deep-Sesr \cite{islam2020simultaneous}, UIECL \cite{li2022beyond}, U-Shape \cite{peng2023u}, TACL \cite{liu2022twin}, Semi-UIR\cite{huang2023contrastive},  Spectroformer\cite{khan2024spectroformer} \textit{etc.} methods of underwater image enhancement.

\begin{table*}[t]
\centering
\setlength{\tabcolsep}{2pt}
\caption{Comparing the proposed method (Ours) with existing state-of-the-art methods on the real-world U45 dataset for underwater image enhancement in terms of quantitative metrics (\textbf{bold} and \underline{underline} indicate the \textbf{best} and \underline{second best} results).}
\label{tab:U45}
\begin{tabular}{l|c|c|c|c|c|c|c|l}
\hline 
\rowcolor{gray!20}
Method & RGHS &WaterNet&UIECL&U-shape&TACL&Semi-UIR& Spectroformer &Ours\\
\hline
UIQM $\uparrow$    & 2.506 & 3.53& 3.59&3.63&4.05&\underline{4.30}&3.243 &\textbf{4.491}\\ 
UISM $\uparrow$    & 5.558 & 6.187 &5.988&5.567&6.698& 7.14&\textbf{7.354} &\underline{7.532}\\  
NIQE $\downarrow$   & 3.872 & 4.596 &3.873&4.309&3.992&\textbf{3.767}&3.842&\underline{3.811}\\ 
BRISQUE $\downarrow$  &\underline{18.51} & 21.15 &20.61&21.56&20.08&23.02 &19.957&\textbf{18.08}\\ 
Time(image/sec) $\downarrow$ & 0.67& 0.370 &0.050&0.079 &0.059&0.044 &\underline{0.040}&\textbf{0.034}\\ 
\hline
\end{tabular}
\end{table*}

\subsection{\textbf{Analysis on Synthetic Datasets}} 
\vspace{-2mm}
We quantitatively compare the proposed method with SOTA techniques using metrics like peak signal-to-noise ratio (PSNR), structural similarity index measure (SSIM), and a non-reference metric of underwater image quality measurement (UIQM). The results for UIEB and UFO-120 datasets can be found in Table \ref{tab:uieb}, and qualitative outcomes are depicted in Figure \ref{fig:uieb_result}. Our method's performance aligns competitively with existing approaches.
\vspace{-3mm}
\subsection{\textbf{Analysis on Real-world Dataset}} 
\vspace{-2mm}
To assess the effectiveness of the proposed method in real-world scenarios, we present results obtained from the U45 dataset. Our quantitative analysis includes metrics such as UIQM (Underwater Image Quality Measure), UISM (Underwater Image Sharpness Measure), NIQE (Naturalness Image Quality Evaluator), and BRISQUE (Blind/Referenceless Image Spatial Quality Evaluator). The tabulated results can be found in Table \ref{tab:U45}. Moreover, qualitative outcomes for the U45, UCCS, and SQUID datasets are showcased in Figure \ref{fig:euvp_result}. The results illustrate that the proposed method significantly improves color balance and visibility in the enhanced images, primarily due to the introduced phase-based mechanisms \textit{Additional quantitative and qualitative outcomes are provided in the supplementary material.}




\begin{table}
\centering
\caption{Computation complexity (lower is better) analysis of proposed method with existing methods.}
\setlength{\tabcolsep}{0.9pt}
\begin{tabular}{l|c|c|c}
\hline
\rowcolor{gray!20}
Method   &Publication&   \#Param$ (\times 10^6$)  & FLOPs ($\times 10^9$)\\
\hline
WaterNet &TIP-19 & 24.81  &193.7 \\
FGAN  &RA-L-20& 7.019  &\textbf{10.23}  \\
Deep-Sesr   &RSS-20& 2.40 & 447.0\\
U-shape   &TIP-23 &65.6 &66.20  \\
UIECL  &TCSVT-22&13.3 & 31.00  \\
TACL  &TIP-22& 11.37& 56.80 \\
Semi-UIR &CVPR-23&\textbf{1.67} &36.43\\
Spectroformer &WACV-24&2.40 &15.75\\
\hline
Ours &--& \underline{1.77}  &\underline{13.0} \\ 
\hline
\end{tabular}
\label{tab:complexity_analysis}
\end{table}

\subsection{\textbf{Computational Complexity Analysis}} 
\vspace{-2mm}
Here, we analyze the computational complexity of the proposed method in comparison with the existing SOTA methods, taking into account the number of trainable parameters, floating-point operations (FLOPs), and run time (in seconds per image). As indicated in Table \ref{tab:complexity_analysis}, the comparison demonstrates that the proposed method has considerably lower computational complexity than current SOTA methods for UIR.



\section{Ablation Study}
\vspace{-3mm}
Following ablation studies are performed on the UIEB \cite{li2019underwater} dataset for evaluating the effectiveness of each proposed component and the various losses.
\vspace{-3mm}
\begin{table}[h]
    \centering
    \caption{Quantitative results comparison of various network settings and losses optimization. \textit{Note: Self-A: Multihead Self Attention, DW- Dynamic Weights, WDW: without DW, PBSA- phase-based self attention, OA- Optimized Attention (without phase), OPA- Optimized Attention (with phase).}}
    \label{tab:ablation}
    \begin{tabular}{c|c|c|c|c|c|c}
        \hline
        \rowcolor{gray!20}{\textbf{Self-A}} & {\textbf{PBSA}} & {\textbf{DW}} & {\textbf{OPA}} & {\textbf{OA}} & \textbf{PSNR} & \textbf{SSIM} \\ \hline
       { $\checkmark$ } &$\times$ &{ $\checkmark$ } &$\times$ & $\times$  &  22.51 & 0.862 \\ 
        $\times$ &{ $\checkmark$ } & { $\checkmark$ } & $\times$ & $\times$ & 24.24 & 0.891 \\ 
        $\times$ &{ $\checkmark$ } &$\times$  &$\times$ & { $\checkmark$ }  &24.46 &  0.901 \\ 
         $\times$ &{ $\checkmark$ }&$\times$ &{ $\checkmark$ } &$\times$ & 24.35&   0.896 \\ \hline
         
      $\times$&{ $\checkmark$ } & { $\checkmark$ }&{ $\checkmark$ } & $\times$ &  \textbf{25.98} & \textbf{0.928} \\ \hline
    \end{tabular}
    \vspace{-4mm}
\end{table}


\subsection{\textbf{Effectiveness of the phase-based self-attention in transformer}}
\vspace{-3mm}
{As the phase of the images is less affected by degradations, the proposed phase-based self-attention mechanism uses phase information to generate an effective attention map for better restoration.} 
To verify this, we tested our hypothesis by conducting experiments with and without phase-based self-attention mechanism. The results, summarized in Table \ref{tab:ablation}, strongly indicate that the phase-based self-attention mechanism contributes to notable improvements
\vspace{-3mm}
\subsection{\textbf{Effectiveness of the optimized phase attention block in feature propagation}}
\vspace{-3mm} Key features are propagated from encoder levels to the decoder with the incorporation of the proposed optimized phase attention block. We evaluated its influence through experiments by comparing the network's performance with and without this block. The outcomes shown in Table \ref{tab:ablation} validate the restored performance achieved when using the optimized phase attention block.

\subsection{\textbf{Effectiveness of the dynamic weights in optimizing the network losses}}
\vspace{-3mm} Assigning proper weight to each loss plays crucial role, whereas manually assigning the weight to each is a tedious task. Assigning learnable weights for losses may solve this problem. As seen from Table \ref{tab:ablation}, we can verify that the performance with the learnable weights for each loss is better as compared to fixed weights.

\begin{figure}[htbp]
\centering   \includegraphics[width=1\linewidth]{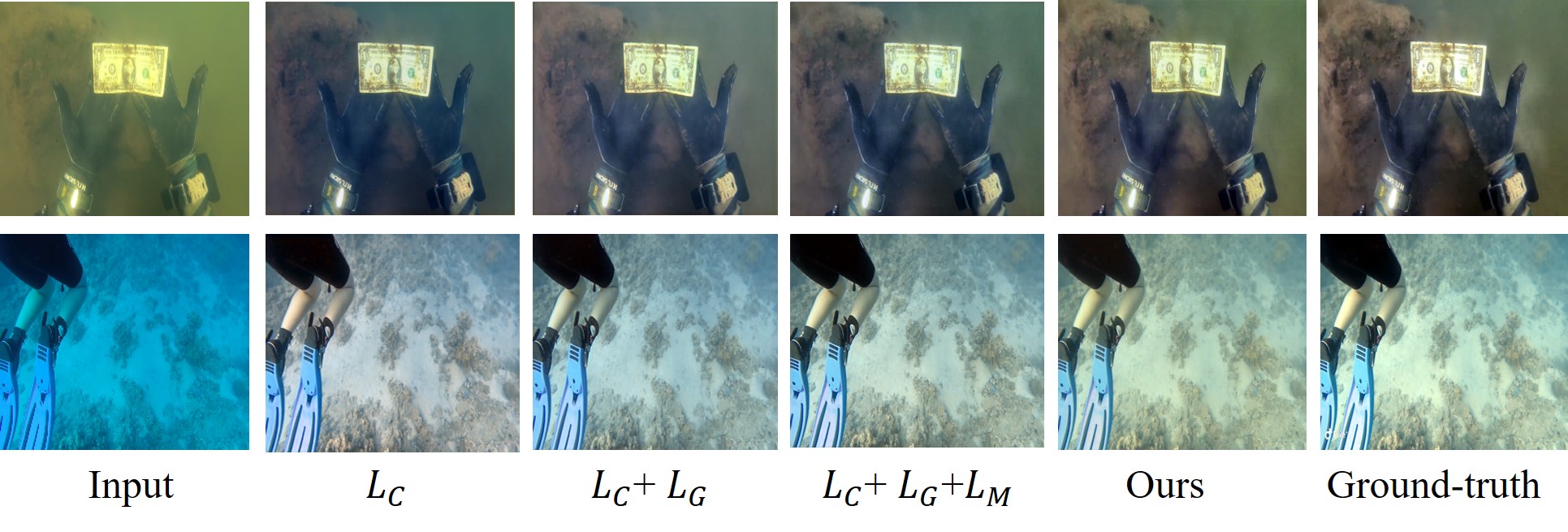}
   \vspace{-7mm}
  \caption{Qualitative comparison of results obtained with various loss settings.}
  \vspace{-4mm}
\label{fig:losses}
\end{figure}

\begin{table}[h]
    \centering
    \caption{Quantitative results comparison of results obtained with various loss settings on UIEB dataset \textit{($L_{C}$: Charbonnier loss, $L_{G}$: Gradient loss,  $L_{M}$: Multiscale Structural Similarity Index (MS-SSIM) loss, and $L_{P}$: Perceptual loss}).}
    \label{tab:loss_ablation}
    \begin{tabular}{ccccc|c|c}
        \hline
       \rowcolor{gray!20} \textbf{Losses} &\textbf{$L_{C}$} & {$L_{G}$} & {$L_{M}$} & {$L_{P}$}  & \textbf{PSNR} & \textbf{SSIM} \\ \hline
       &{ $\checkmark$ } & & &  &  23.97 &0.890 \\ 
        &$\checkmark$ &{ $\checkmark$ } & {$\times$  } & $\times$ & 24.15   &  0.907 \\ 
       & $\checkmark$ &{ $\checkmark$ } &$\checkmark$ &$\times$ & 25.11  &  0.912\\ \hline
      \textbf{Ours}&$\checkmark$& $\checkmark$ & $\checkmark$ & $\checkmark$&  \textbf{25.98} & \textbf{0.928} \\ \hline
    \end{tabular}
\end{table}

\subsection{\textbf{Effectiveness of various loss settings}}
\vspace{-3mm}
We also presented the impact of various employed losses on underwater image enhancement, both quantitatively in Table \ref{tab:loss_ablation} and qualitatively in Figure \ref{fig:losses}. This comparison shows the effectiveness of utilizing the combination of $L_C, L_G, L_M$, and $L_P$.
\vspace{-2mm}
\begin{figure}[htbp]
\begin{center}  \includegraphics[width=1\linewidth,height=0.5\linewidth]{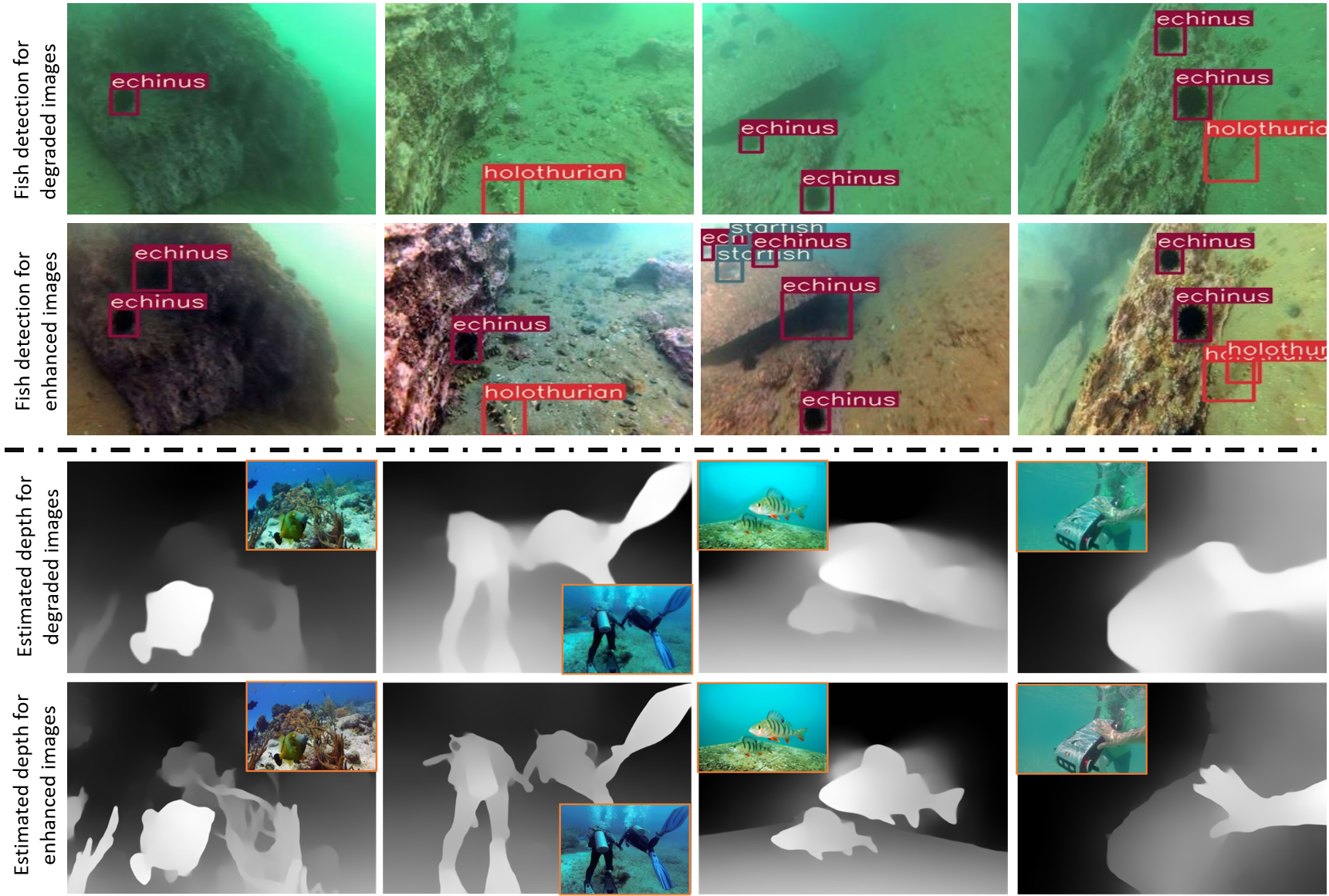}
\end{center}
\vspace{-2mm}
   \caption{Comparison of degraded images and corresponding enhanced images for fish detection (\textit{upper part}) 
 and depth estimation (\textit{bottom part}) applications.}
   \vspace{-4mm}
\label{fig:depth}
\end{figure}
\section{Applications of the Proposed Method}
\vspace{-3mm}
In this section, we explore the practicality of the proposed method for various tasks discussed below.
\subsection{\textbf{Down stream applications }}
\vspace{-2mm}Underwater image restoration as a preprocessing step improves object detection and depth estimation tasks' performance. Our method shows superior performance on YOLO-v3 \cite{redmon2018yolov3} for object detection and DPT \cite{Ranftl2021} for depth estimation, as illustrated in Figure \ref{fig:depth}. The results confirm the applicability of our approach for improving the performance of higher-level computer vision.
\vspace{-2mm}
\begin{table}
\caption{Quantitative results comparison of state-of-the-art (TFFR \cite{Ma_2022_CVPR}, MIRNetV2 \cite{Zamir2022MIRNetv2}) and the proposed method (Ours) on real world dataset provided in \cite{Chen2018Retinex}.}
\label{tab:lowlight_quant}
\centering
\begin{tabular}{l|c|c}
\hline
\rowcolor{gray!20} Methods              & NIQE $\downarrow$  & BRISQUE $\downarrow$  \\
\hline
TFFR \cite{Ma_2022_CVPR}        & 5.620 & 29.652 \\
MIRNetV2 \cite{Zamir2022MIRNetv2} & 7.513 & 33.929 \\ \hline
Ours  & \textbf{4.535} & \textbf{22.117}     \\
\hline
\end{tabular}
\vspace{-3mm}
\end{table}
\begin{figure}[htbp]
\begin{center}
  \includegraphics[width=1\linewidth,height=0.4\linewidth]{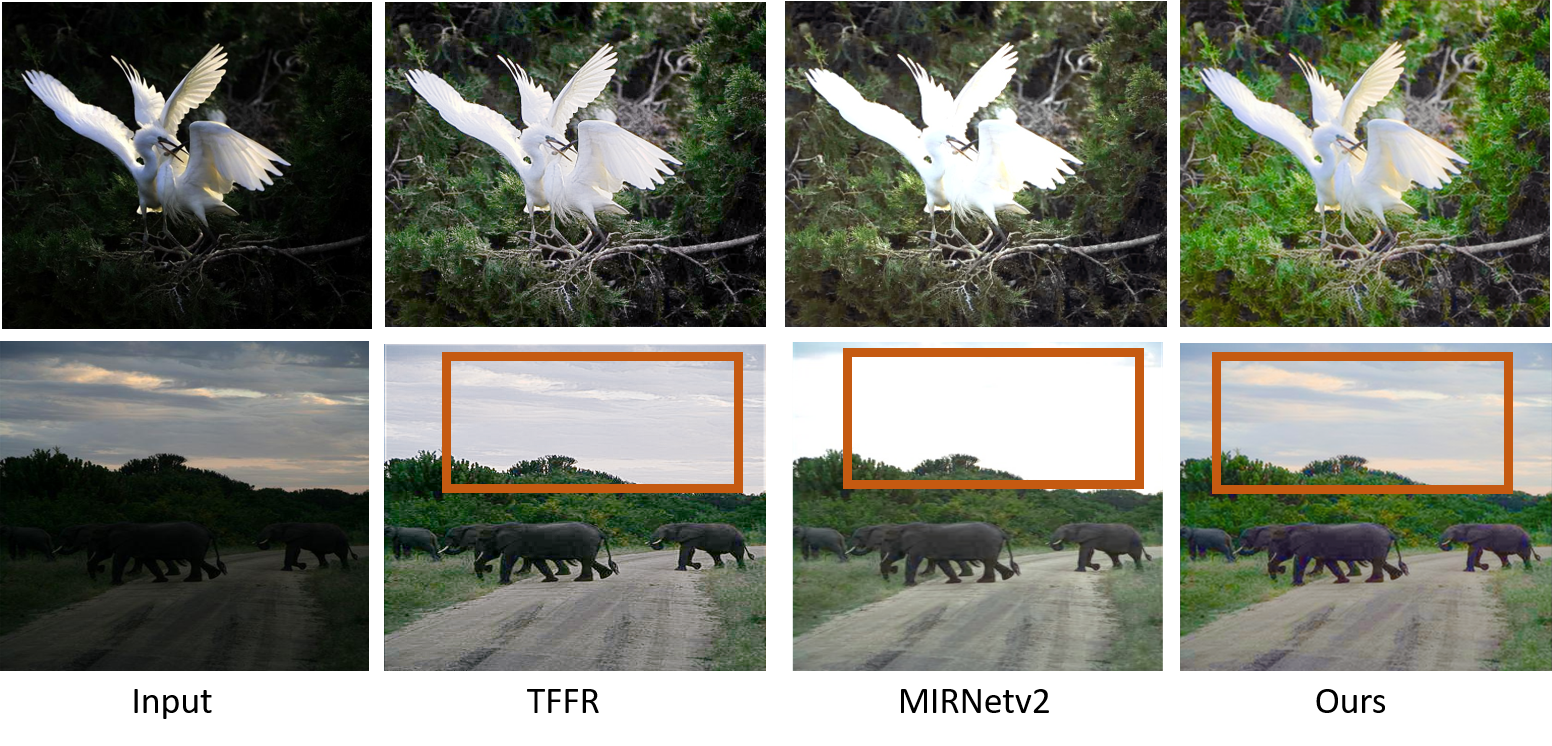}
\end{center}
\vspace{-6mm}
   \caption{Qualitative comparison of the proposed method (Ours) with existing SOTA methods (TFFR \cite{Ma_2022_CVPR}, MIRNetV2 \cite{Zamir2022MIRNetv2}) for real-world low light image enhancement.}
   \vspace{-3mm}
\label{fig:real-world LL}
\end{figure}
\subsection{\textbf{Low-light image enhancement}}
\vspace{-2mm}Low-light image enhancement is another important task that focuses on improving visibility in low-light regions while preserving color consistency and reducing noise. To demonstrate the applicability of our proposed method, we train the network on 970 augmented synthetic image pairs (low-light and ground truth) from \cite{RETINEX}. Our method's performance is showcased through qualitative results depicted in Figure \ref{fig:real-world LL}, and quantitative evaluations summarized in Table \ref{tab:lowlight_quant}, specifically tailored for real-world low-light image enhancement tasks. These results demonstrate the superiority of our proposed method over state-of-the-art techniques in this domain.
 \textit{Additional, qualitative results are available in the supplementary material}.
\section{Failure Case}
\begin{figure}[htbp]
\begin{center}
  \includegraphics[width=1\linewidth]{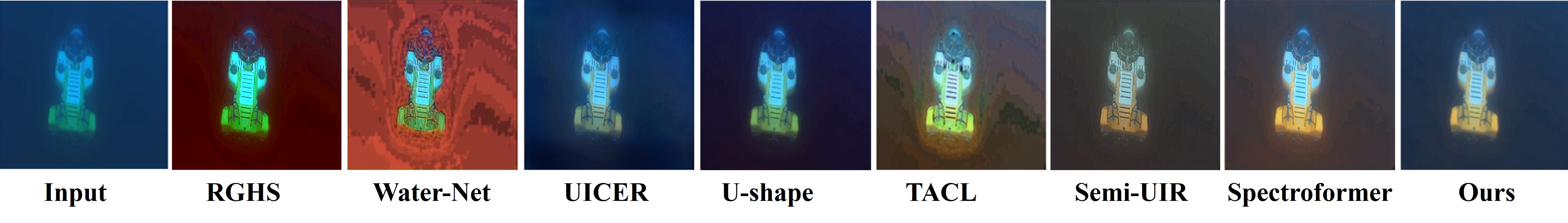}
\end{center}
\vspace{-4mm}
   \caption{Qualitative results analysis with the proposed (ours) and state-of-the-art methods on the real-world C-60 UIEB dataset for underwater image restoration.}
\label{fig:falure}
\end{figure}

\vspace{-2mm}
Analysis of Figure  \ref{fig:falure} reveals the challenges faced by existing methods in enhancing muddy and blurry underwater images. Despite these difficulties, our approach consistently surpasses current state-of-the-art methods in underwater image enhancement. However, it's important to note that there remains room for improvement in our outputs, particularly in terms of deblurring. We still need to improve the deblurring aspect of our output, which we are considering for the future scope.
\vspace{-5mm}
\section{Conclusion}
\vspace{-2mm}
This paper introduces a novel lightweight phase-based transformer network for underwater image restoration. The network utilizes a phase-based attention mechanism to extract non-contaminated features effectively, and an optimized phase attention block ensures proper structural reconstruction and prominent feature propagation. Learnable parameters are used to weight the losses during training. Extensive evaluations on synthetic and real-world datasets, along with ablation studies, validate our method's effectiveness. Its applicability also extends to diverse areas like low-light enhancement, enhanced object detection, and improved depth estimation.

\section*{Acknowledgement}
\vspace{-2mm}
This research was supported by MoES (Grant MoES/PAMC/DOM/04/2022 - E-12710), TIHIITG202204, and (d-real) Science Foundation Ireland (Grant 18/CRT/6224). The authors thank the CVPR Lab members at Trinity College Dublin and IIT Ropar for their support.

\bibliographystyle{splncs04}
\bibliography{egbib}
\end{document}